\documentclass[10pt,twocolumn,letterpaper]{article}
\usepackage[pagenumbers]{cvpr} 

\definecolor{cvprblue}{rgb}{0.21,0.49,0.74}
\usepackage[pagebackref,breaklinks,colorlinks,allcolors=cvprblue]{hyperref}

\def\confName{CVPR}
\def\confYear{2027}

\newtheorem{fdpotheorem}{Theorem}

\newtheorem{fdpoproposition}[fdpotheorem]{Proposition}

\AtBeginDocument{
    \crefname{fdpotheorem}{theorem}{theorems}
    \Crefname{fdpotheorem}{Theorem}{Theorems}
    \crefname{fdpolemma}{lemma}{lemmas}
    \Crefname{fdpolemma}{Lemma}{Lemmas}
    \crefname{fdpoproposition}{proposition}{propositions}
    \Crefname{fdpoproposition}{Proposition}{Propositions}
}

\title{DiPACT: Distributional Preference Alignment Approach to \\ 
Continual Quantum Material Characterization}

\author{
Sankalp Pandey\textsuperscript{1},
Hoang-Quan Nguyen\textsuperscript{1},
Thanh-Dat Truong\textsuperscript{1},
Tim Faltermeier\textsuperscript{2},\\
Taylor D. Sparks\textsuperscript{3},
Mike Kirby\textsuperscript{3},
Nicholas J. Borys\textsuperscript{3},\\
Hugh Churchill\textsuperscript{4},
Khoa Luu\textsuperscript{1}
\\[0.4em]
\textsuperscript{1}CVIU Lab, University of Arkansas, USA
\qquad
\textsuperscript{2}Montana State University, USA\\
\textsuperscript{3}University of Utah, USA
\qquad
\textsuperscript{4}Department of Physics, University of Arkansas, USA
\\[0.4em]
{\ttfamily\small
\textsuperscript{1}\{sankalpp,hn016,tt032,khoaluu\}@uark.edu}\\
{\ttfamily\small
\textsuperscript{2}timfaltermeier@montana.edu}\\
{\ttfamily\small
\textsuperscript{3}\{taylor.sparks,mike.kirby,nicholas.borys\}@utah.edu}\\
{\ttfamily\small
\textsuperscript{4}hchurch@uark.edu}
\\[0.3em]
{\small\url{https://uark-cviu.github.io/projects/DiPACT}}
}

\begin{document}
\maketitle
\begin{abstract}
Vision-language models enable instruction-guided localization and recognition of two-dimensional quantum material flakes in microscopy images. Characterizing the atomic layer counts of these flakes is essential for selecting candidates with the desired properties for next-generation device fabrication, but this identification is challenged by similar optical signatures across layer classes and substantial appearance variation within each class.
As new materials are introduced, models must learn these subtle visual distinctions while retaining earlier knowledge.
Conventional Direct Preference Optimization (DPO) ranks correct and incorrect characterizations individually, without aggregating the same visual distinction across images.
We introduce Distributional Preference Alignment for Continual Training (DiPACT), which groups preferences sharing a directed correction and optimizes a smooth minimum of their margins.
The objective emphasizes weak comparisons and aligns each group with a frozen reference selected for earlier or newly introduced classes.
Our analysis connects the objective to Fisher discrimination under linear-Gaussian assumptions and bounds the loss of previously correct preference rankings under comparison coverage.
We also present the Continual Material Annotated Dataset (CoMAt), combining 205,362 expert-reviewed in-house annotations with public microscopy data for layer- and material-incremental evaluation.
Experiments demonstrate state-of-the-art continual learning and fine-grained characterization performance among the evaluated methods, with benefits extending to visual recognition beyond the materials domain.\footnote{Code, pretrained models, and data will be released.}
\end{abstract}
\vspace{-2mm}
\section{Introduction}
\label{sec:intro}

\begin{figure}[t]
\centering
\includegraphics[width=\linewidth]{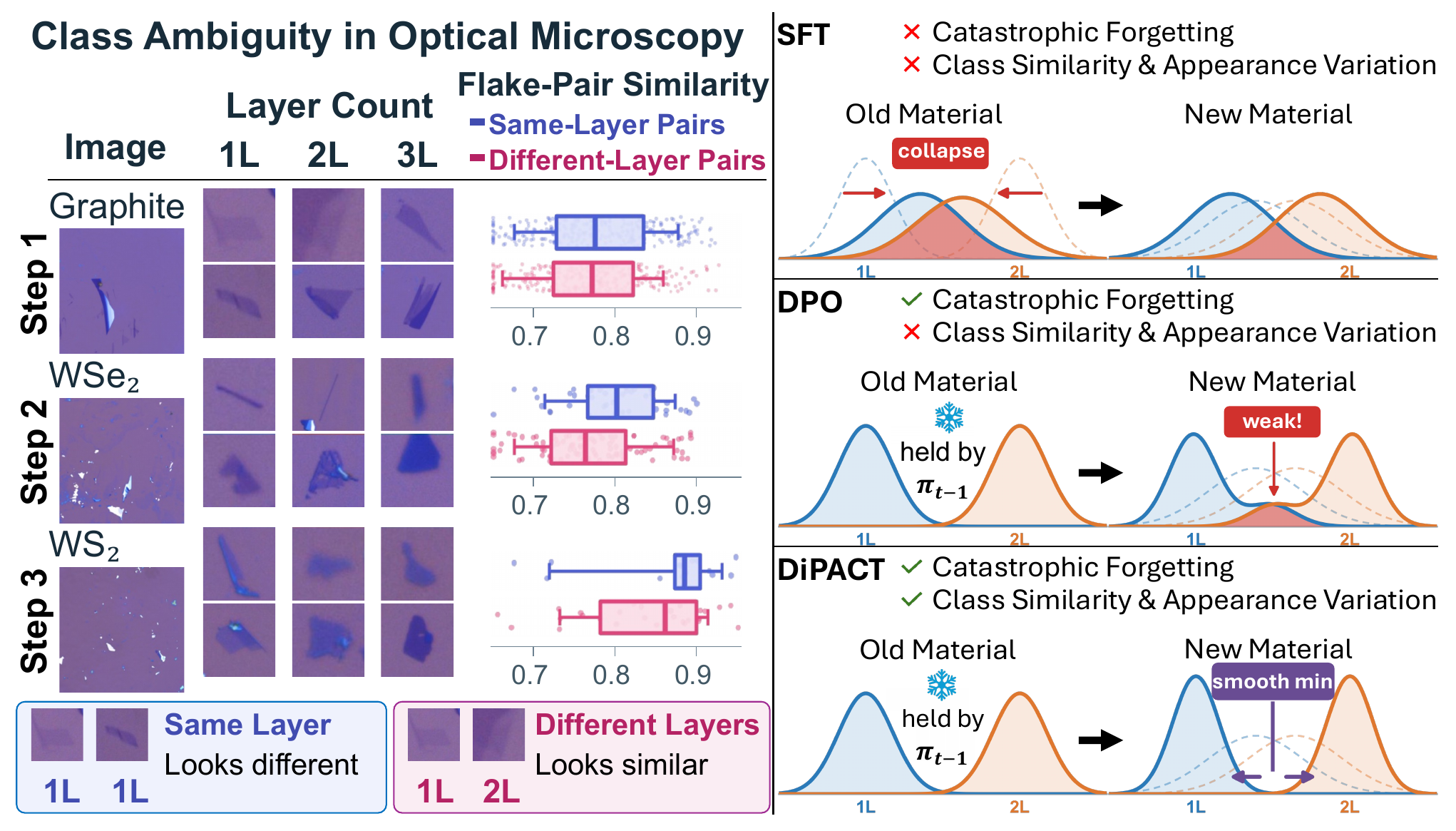}
\caption{\textbf{Our DiPACT approach to Continual Quantum Material Characterization.}
Similarities between layer classes and appearance variation within them challenge continual characterization. The schematic contrasts forgetting under sequential SFT with unresolved distinctions under DPO.
\textbf{DiPACT} groups corrections across images and emphasizes weak preference margins to improve discrimination and retention \textbf{(Best viewed in color)}.}
\label{fig:teaser}
\vspace{-6mm}
\end{figure}

Foundation vision-language models (VLMs) have shown promising capabilities in microscopy image understanding and material characterization~\cite{nguyen2026qupaint}.
In two-dimensional (2D) material characterization, these capabilities connect the appearance of individual material fragments, or flakes, to their atomic layer counts, supporting sample selection for scientific study and device fabrication.
As additional materials and layer categories enter this workflow, models must learn new visual distinctions while preserving what they learned previously.
This evolving recognition problem motivates continual adaptation of grounded vision-language models.

\vspace{1mm}
\noindent\textbf{Challenges in Continual Grounded Recognition.}
Continual adaptation must accommodate both ambiguity between classes and variation within them.
Different layer counts can have similar optical signatures, while the appearance of flakes with the same thickness may vary~\cite {blake2007making}.
An update must therefore learn subtle distinctions across diverse appearances without disrupting those established previously (\Cref{fig:teaser}).
CLIFF~\cite{pandey2025cliff} addresses continual flake classification through material-specific prompts and residual heads, using stored examples and distillation to retain earlier knowledge.
Its classification formulation does not jointly refine flake localization and layer assignments in structured predictions.
We study this grounded setting using QuPAINT~\cite{nguyen2026qupaint}, with material context supplied in the instruction.
Since our approach uses no replay, the learning objective must make effective use of the visual decisions supported by these observations.
This motivates comparing correct and incorrect characterizations, and learning the same correction reliably across images.

\vspace{1mm}
\noindent\textbf{Preference Learning for Fine-Grained Recognition.}
These structured outputs allow supervision to isolate a specific visual decision. Reviewed characterizations can be paired with alternatives containing controlled errors.
Supervised fine-tuning (SFT) learns to reproduce the reviewed response, while Direct Preference Optimization (DPO) explicitly ranks it above the alternative relative to a frozen reference~\cite{rafailov2023direct}.
For continual multimodal learning, $\phi$-DPO \cite{truong2026faidpo} introduces previous-checkpoint references and focal weighting of individual preferences.
However, the same correction can be easy in one image and unreliable in another.
For example, preferring correct bilayer over incorrect trilayer graphite expresses the same directed distinction across flakes with different appearances.
This suggests optimizing the reliability of that shared correction across images, with particular attention to its weakest instances.

\vspace{1mm}
\noindent\textbf{Distributional Preference Alignment.}
We present the new \textbf{\textit{Distributional Preference Alignment for Continual Training (DiPACT) approach}} to optimize these shared corrections at the group level. DiPACT groups instances of the same directed correction and aggregates their correct-versus-incorrect margins with a smooth minimum.
The resulting statistic is compared with a matched aggregate from a frozen reference, giving one alignment objective for the group.
Smaller current margins receive larger relative gradient weights, directing updates toward images where the shared distinction remains weak.
This aggregation explicitly couples comparable corrections within an objective, extending beyond focal weighting of individual preference pairs.
Under linear-Gaussian assumptions, we connect the objective to Fisher discrimination, relating class separation to within-class variability.

\vspace{1mm}
\noindent\textbf{Continual Adaptation with Knowledge Retention.}
Continual characterization must accommodate new visual distinctions without eroding those established in earlier rounds.
DiPACT uses class-routed references to guide these updates, anchoring established preferences while learning corrections involving new classes.
Our analysis establishes a conditional bound on preference reversals when current comparisons cover the earlier ones, linking the alignment objective to knowledge retention.

In addition, a new \textbf{\textit{Continual Material Annotated (CoMAt) quantum materials dataset}} is introduced to support continual grounded recognition. 
CoMAt provides a new in-house microscopy collection with 205,362 expert-reviewed annotations across five materials, including flake bounding boxes, layer labels, and non-flake annotations.
We supplement this collection with the public MoS$_2$ and WTe$_2$ dataset~\cite{masubuchi2020deep} to extend evaluation across materials and acquisition sources.
Together, these data support layer-incremental and material-incremental protocols, while the flake annotations enable structured preferences targeting classification, detection, and localization errors.

\vspace{1mm}
\noindent\textbf{Contributions of this Work.} There are three main contributions in this work. 
First, we introduce DiPACT, a group-level preference objective that emphasizes weak visual distinctions and uses class-routed references for continual grounded recognition.
Our analysis connects the objective to Fisher discrimination and establishes a conditional preference-retention bound.
Second, we present CoMAt, a large-scale microscopy dataset with structured preference supervision and layer- and material-incremental protocols.
Finally, extensive experiments and ablation studies demonstrate DiPACT's effectiveness and state-of-the-art (SOTA) performance in continual learning and material characterization, with benefits extending to visual recognition beyond the materials domain.

\vspace{-2mm}
\section{Related Work}

\noindent\textbf{Two-Dimensional Material Characterization.}
Automated microscopy characterization has progressed from UNet segmentation~\cite{ronneberger2015u,han2020deep} and Mask R-CNN detection~\cite{he2017mask,masubuchi2020deep} to self-attention with soft labels~\cite{nguyen2024two} and methods addressing false negatives~\cite{luu2024automatically}.
Gaussian mixture models provide an alternative based on optical appearance~\cite{uslu2024open}.
Recent work has explored high-throughput identification~\cite{crimmann2025high}, material foundation models~\cite{uslu2025maskterial}, general segmentation models~\cite{kirillov2023segment,ravi2024sam}, and integration with fabrication and sample-processing workflows~\cite{mckenzie2024fabrication,Ouaj_2024,Ouaj_2023,jackering2025super,courtney2025automated}.
QuPAINT extends characterization to instruction-guided localization and layer recognition~\cite{nguyen2026qupaint}, while agentic systems combine VLMs with tools for interactive scientific analysis~\cite{du2026eaa}. These approaches help support the study of quantum applications and algorithms \cite{nguyen2026quacod, nguyen2025qmoe, nguyen2024qclusformer,holliday2025quadro,nguyen2024quantum,nguyen2024hierarchical,nguyen2025diffusion, holliday2026advanced, holliday2024hybrid}.
These advances motivate continual adaptation as the recognized materials and layer categories expand.

\noindent\textbf{Continual Learning.}
Continual learning aims to adapt models to sequential tasks while mitigating catastrophic forgetting. X\cite{truong2023fredom,nguyen2026cobra, truong2023fairness,truong2024conda,truong2025falcon}. Existing approaches include rehearsal-based \cite{buzzega2020dark}, regularization-based \cite{kirkpatrick2017overcoming}, architecture-based \cite{douillard2022dytox}, and prompt-based methods \cite{wang2022learning}. Recent studies have extended continual learning to large multimodal models (LMMs). Chen et al. \cite{chen2024coin} introduced a benchmark for continual instruction tuning, while Zeng et al. \cite{zeng2025modalprompt} explored dual-modality guided prompting. Other approaches investigate continual instruction tuning \cite{cao2024continual,guo2025hide}, forgetting mitigation \cite{chen2025sefe,zhao2025mllm}, and sparse memory fine-tuning \cite{lin2025continual}. Fairness under imbalanced data distributions has been explored in continual semantic segmentation \cite{truong2023fairness,truong2025falcon}, with more recent work extending this direction to large multimodal models through preference-based learning \cite{truong2026faidpo}.
However, reference-based alignment of individual preferences does not explicitly account for how reliably the same distinction is made across observations.
DiPACT groups comparable corrections across images to emphasize weak material-layer discrimination, while reference anchors guide retention of earlier distinctions.

\vspace{-2mm}
\section{The CoMAt Dataset and Benchmark}
\label{sec:dataset}
This work introduces the Continual Material Annotated Dataset to study how models learn to characterize additional quantum materials and layer counts while retaining previously learned distinctions.
CoMAt combines a new collection of expert-reviewed microscopy images with continual-learning protocols and preference construction from reviewed annotations.
The in-house collection covers graphite, WS$_2$, WSe$_2$, hBN, and MoS$_2$.
To extend evaluation across quantum materials and acquisition sources, the material-incremental benchmark incorporates public MoS$_2$ and WTe$_2$ images from Masubuchi et al.~\cite{masubuchi2020deep}.
The new in-house annotations and the public benchmark extension are reported separately.

\begin{figure}[t!]
\centering
\includegraphics[width=1\linewidth]{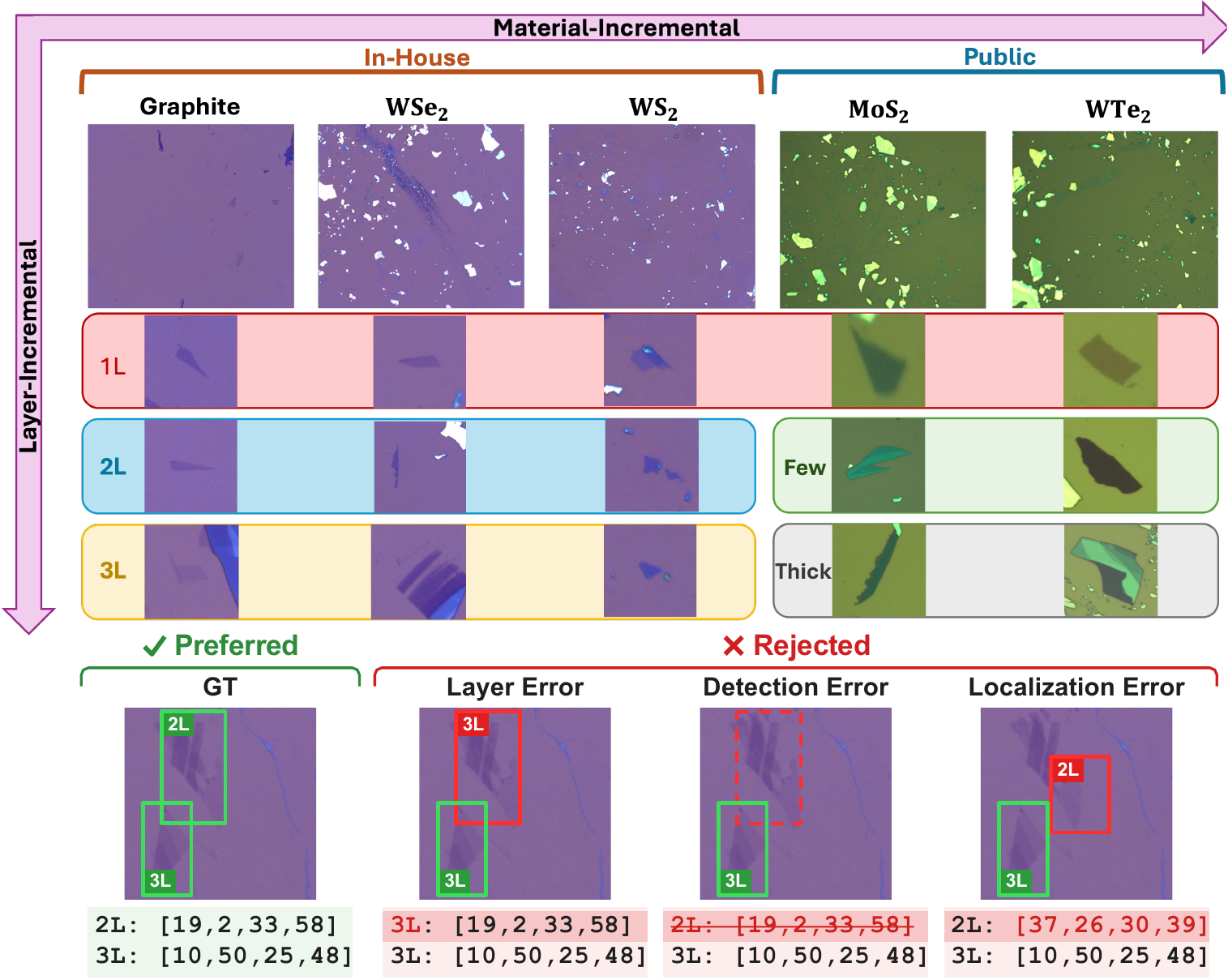}
\caption{\textbf{CoMAt Dataset and Preference Construction.}  Microscopy images showcasing material diversity and layer categories across in-house and public data. Preferred responses follow ground-truth annotations. Rejected responses introduce a layer-label error, missed detection, or localization error while preserving the remaining annotations \textbf{(Best viewed in color)}.}
\label{fig:dataset_protocols}
\end{figure}

\subsection{Data Collection and Annotation}
\label{sec:dataset_annotation}
The in-house images are collected during characterization of exfoliated 2D quantum materials.
\textbf{\textit{The acquisition pipeline registers candidate flakes for review by materials science experts}} with experience in 2D-quantum material characterization.
These reviewers assign monolayer, bilayer, trilayer, thick, or non-flake (artifact) labels using the expected color and contrast relative to the substrate for the corresponding material and imaging conditions. The in-house assessment procedure uses higher-magnification and dark-field inspection to verify assignments when they are uncertain. These reviewers correct or add bounding boxes and descriptions where appropriate.
Each flake retains its review status, allowing verified targets to be distinguished from unlabeled candidates.


\subsection{Dataset Statistics}
\label{sec:dataset_statistics}

The proposed in-house quantum materials collection contains 205,362 expert-reviewed
annotations: 153,849 flakes and 51,513 non-flake artifacts.
\Cref{tab:dataset_labels} provides the per-material and
layer-category breakdown. 

\begin{table}[t]
\centering
\caption{\textbf{In-House Annotation Breakdown.}
Reviewed annotations by material and layer category.
1L, 2L, and 3L denote monolayer, bilayer, and trilayer.
Neg. denotes non-flake artifacts.}
\label{tab:dataset_labels}
\footnotesize
\setlength{\tabcolsep}{0pt}
\begin{tabular*}{\linewidth}{@{\extracolsep{\fill}}lrrrrr@{}}
\toprule
Material & 1L & 2L & 3L & Thick & Neg. \\
\midrule
Graphite & 978 & 1,082 & 1,163 & 113,970 & 20,261 \\
WS$_2$ & 5 & 10 & 10 & 12,876 & 15,692 \\
WSe$_2$ & 11 & 4 & 14 & 22,238 & 4,158 \\
hBN & 0 & 1 & 23 & 617 & 4,118 \\
MoS$_2$ & 0 & 2 & 3 & 842 & 7,284 \\
\midrule
\textbf{Total} & \textbf{994} & \textbf{1,099} & \textbf{1,213} & \textbf{150,543} & \textbf{51,513} \\
\bottomrule
\end{tabular*}
\end{table}


The material-incremental protocol uses 26,665 supervised
flakes, comprising 25,181 in-house annotations and
1,484 public Masubuchi annotations.
Of these, 25,102 are used for training and 1,563 for validation.
\Cref{tab:dataset_material_rounds} provides the per-material
breakdown. Public few-layer labels are retained without assigning
exact bilayer or trilayer counts.

\subsection{Continual Evaluation and Preference Construction}
\label{sec:dataset_protocol}
\Cref{fig:dataset_protocols} shows representative microscopy
images, layer categories, and annotation-based preference construction.

\vspace{1mm}
\noindent\textbf{Layer-Incremental Characterization.}
Training begins with monolayer and thick graphite flakes, followed by bilayer and trilayer classes in successive rounds. Each round uses a different set of training chips and includes reviewed annotations for all layer classes admitted so far.
Earlier classes may therefore occur in newly acquired images, while historical training images are not replayed.
After each round, evaluation covers all admitted classes, separating performance on newly presented and earlier-layer classes.

\vspace{1mm}
\noindent\textbf{Material-Incremental Characterization.}
Materials are introduced in the order listed in
\Cref{tab:dataset_material_rounds}.
Each round trains only on the current material and evaluates all materials introduced so far.
This setting measures learning and forgetting as both material appearance and acquisition source change.

\vspace{1mm}
\noindent\textbf{Annotation-Based Preferences.}
The reviewed annotations support preferences between complete characterizations of the same image.
A preferred output follows the verified annotations, while an alternative introduces a specified layer-identification, detection, or localization error.
Each comparison retains its material, correction type, and relevant layer labels, allowing comparable decisions to be grouped across images and assigned to their learning rounds.
Preferences can be constructed during training from the annotations, rather than requiring experts to rank every output pair separately.
The preference families used in each experiment are specified in \Cref{sec:exp}.

\vspace{1mm}
\noindent\textbf{Data Separation.}
In-house training and validation images are separated by acquisition chip.
For Masubuchi et al.~\cite{masubuchi2020deep}, we retain the published training and validation split and remove training images that are byte-identical to validation images.
We do not claim chip-disjoint splits for the public data, whose acquisition-chip identities are unavailable.
Unverified annotations are excluded from supervised targets and retained as ignore flakes during evaluation.

\begin{table}[t]
\centering
\caption{\textbf{Material-Incremental Annotation Split.}
Flake counts across five rounds.
Each round introduces one material.}
\vspace{-2mm}
\label{tab:dataset_material_rounds}
\footnotesize
\setlength{\tabcolsep}{3pt}
\begin{tabular*}{\linewidth}{@{\extracolsep{\fill}}cllrr@{}}
\toprule
Round & Material & Source & Train & Val. \\
\midrule
0 & Graphite & In-house & 13,653 & 879 \\
1 & WSe$_2$ & In-house & 7,292 & 75 \\
2 & WS$_2$ & In-house & 2,967 & 315 \\
3 & MoS$_2$ & Masubuchi & 615 & 147 \\
4 & WTe$_2$ & Masubuchi & 575 & 147 \\
\midrule
\multicolumn{3}{l}{\textbf{Total}}
& \textbf{25,102} & \textbf{1,563} \\
\bottomrule
\end{tabular*}
\vspace{-4mm}
\end{table}
\section{The Proposed DiPACT Approach}
\label{sec:fdpo_method}
\begin{figure*}[t]
\centering
\includegraphics[width=1\textwidth]{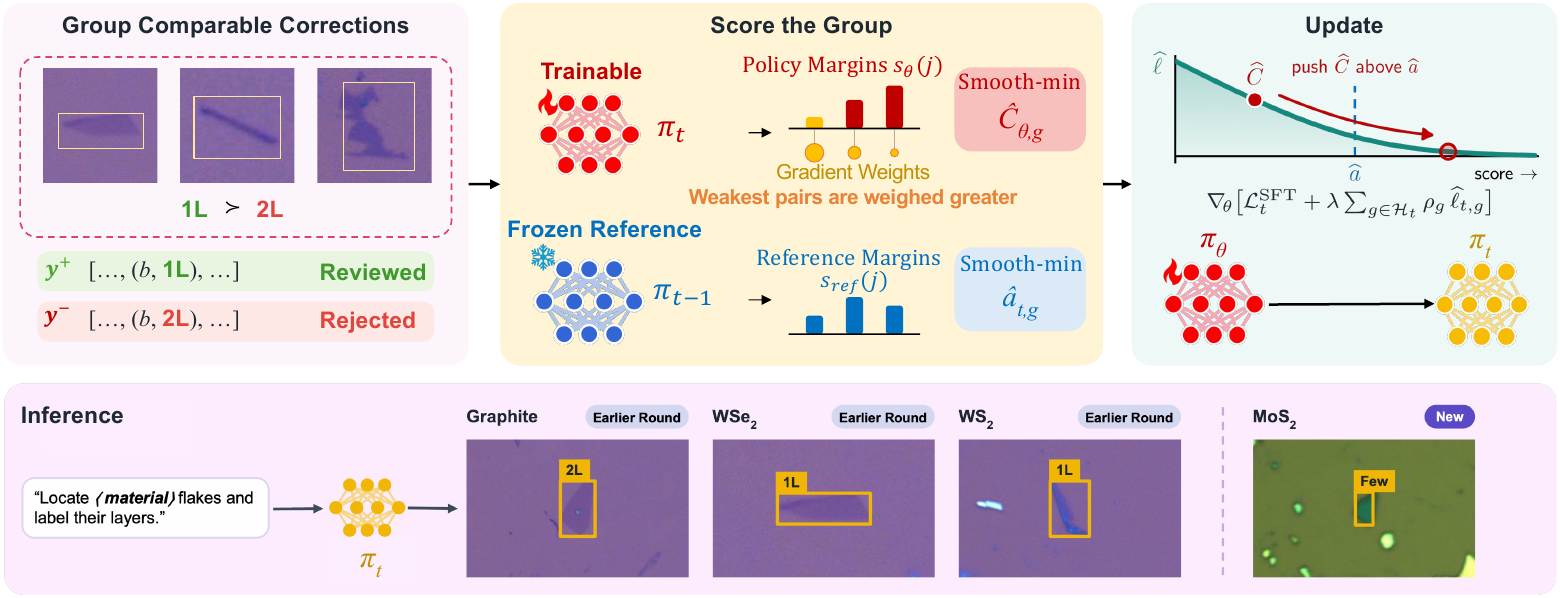}
\vspace{-2mm}
\caption{\textbf{Overview of the proposed DiPACT approach.}
Preference pairs sharing a directed correction form a group across images.
Smooth-minimum aggregation emphasizes low-margin comparisons through larger gradient weights.
The loss compares the trainable model's aggregate with a matched frozen-reference aggregate.
The resulting model $\pi_t$, obtained after training at round $t$, performs grounded characterization of all materials and layer classes introduced so far \textbf{(Best viewed in color)}.}
\label{fig:continual}
\vspace{-4mm}
\end{figure*}

DiPACT learns visual distinctions across images by grouping preferences that express the same directed correction.
Its central objective compares smooth-minimum aggregates of current and reference preference margins, giving greater influence to observations on which that correction remains difficult.
\Cref{fig:continual} illustrates the method, whose discrimination and conditional retention properties we analyze below.

\subsection{Task and Preference Supervision}
\label{sec:fdpo_task}

At round $t$, the recognized material-layer classes expand as $\mathcal C_t=\mathcal C_{t-1}\cup\Delta\mathcal C_t$, where $\Delta\mathcal C_t$ contains the newly introduced classes and $\mathcal C_{-1}=\varnothing$.
Training uses the current microscopy dataset $\mathcal D_t$ without historical-image replay, and evaluation covers all classes introduced so far.

Given an image $I$ and an instruction $q_m$ specifying material $m$, the vision-language model $\pi_\theta$ with parameters $\theta$ generates a response $y$ encoding $[(b_k,\ell_k)]_{k=1}^{K}$.
Here, $b_k$ and $\ell_k$ are the bounding box and layer label of predicted flake $k$, $K$ is the number of predictions, and $(m,\ell_k)$ is its material-layer class.
Boxes encode top-left position, width, and height as percentages of image dimensions.
We write the input as $x=(I,q_m)$ and denote the completed round-$t$ model by $\pi_t$.

\vspace{1mm}
\noindent\textbf{Controlled Preference Pairs.}
\label{sec:fdpo_preferences}
Each preference record $j$ pairs a reviewed response $y_j^+$ with an alternative $y_j^-$ for the same input $x_j$.
A label edit changes a flake's layer assignment while holding its box fixed, isolating a visual class distinction.
Omission, false-positive, and localization edits similarly target missed flakes, unsupported detections, and incorrect boxes.
All unedited output fields remain unchanged, and false-positive edits are used only where review establishes that the added detection is incorrect.
Pairs are constructed from annotation-based edits and reviewed errors in model predictions, with refreshed comparisons exposing errors of the evolving model.

\vspace{1mm}
\noindent\textbf{Preference Margins.}
To compare the responses, we score either the full response or the fields affected by the edit, using the same convention for the trainable and reference models.
For a token mask $M=(M_r)_{r=1}^{|y|}$ with $M_r\in\{0,1\}$, the selected log-likelihood is defined as follows,
\begingroup
\small
\begin{equation}
 L_\theta(y\mid x,M)
 =\sum_{r=1}^{|y|}M_r\log\pi_\theta(y_r\mid x,y_{<r}).
 \label{eq:fdpo_masked_likelihood}
\end{equation}
\endgroup
In \Cref{eq:fdpo_masked_likelihood}, $|y|$ is the response length, $y_r$ is token $r$, and $y_{<r}$ is its preceding response context, including tokens excluded from the score.
With masks $M_j^+$ and $M_j^-$ for the two responses, the preference margin is as follows,
\begingroup
\small
\begin{equation}
 s_\theta(j)=L_\theta(y_j^+\mid x_j,M_j^+)
 -L_\theta(y_j^-\mid x_j,M_j^-).
 \label{eq:fdpo_score}
\end{equation}
\endgroup
Positive values in \Cref{eq:fdpo_score} favor the reviewed characterization.
Mask construction is detailed in the appendix, and experimental scoring configurations in \Cref{sec:exp_setup}.

For comparison, DPO~\cite{rafailov2023direct} optimizes a reference-adjusted margin for each pair:
\begingroup
\small
\begin{equation}
 \mathcal L_t^{\mathrm{DPO}}
 =-\mathbb E_j\log\sigma\!\left(
 \beta[s_\theta(j)-s_{\mathrm{ref},t}(j)]\right).
 \label{eq:fdpo_pairwise_dpo}
\end{equation}
\endgroup
Here, $\mathbb E_j$ averages over sampled records at round $t$, $\sigma(u)=1/(1+e^{-u})$, $\beta>0$ scales the preference logit, and $s_{\mathrm{ref},t}(j)$ is the frozen-reference margin.
\Cref{eq:fdpo_pairwise_dpo} is ordinary DPO with full-response scores and a targeted logistic comparison with edited-field scores.
Although $\phi$-DPO applies focal weighting to individual preferences~\cite{truong2026faidpo}, neither objective aggregates margins for a shared visual correction across images before applying the preference loss.
DiPACT introduces this aggregation to emphasize observations where the same distinction remains weak.

\subsection{Distributional Preference Alignment}
\label{sec:fdpo_distribution}
\noindent\textbf{Grouping Shared Corrections.}
Let $\mathcal G_t$ index the preference groups at round $t$, and let $G_{t,g}$ contain the records in group $g$.
For label edits, each group fixes a material and a directed correction, such as correct bilayer graphite versus incorrect trilayer graphite.
The reverse correction forms a separate group when reviewed examples are available.
Other groups fix the error type and affected material-layer class.
Thus, a group represents one correction across different images and appearances, with each record retaining its own input and candidate responses.

\noindent\textbf{Group-Level Preference Objective.}
We summarize a correction by a smooth minimum of its margins, making the aggregate sensitive to its weakest observations.
For a sampled batch $B_{t,g}\subseteq G_{t,g}$ with $n_g=|B_{t,g}|\ge1$, define
\begingroup
\small
\begin{equation}
 \widehat C_{\theta,g}
 =-\tau\log\left[
 \frac{1}{n_g}\sum_{j\in B_{t,g}}
 \exp\!\left(-\frac{s_\theta(j)}{\tau}\right)\right].
 \label{eq:fdpo_smoothmin}
\end{equation}
\endgroup
The temperature $\tau>0$ controls emphasis on low margins, with \Cref{eq:fdpo_smoothmin} approaching the minimum as $\tau\to0$ and the mean as $\tau\to\infty$.
Hats denote batch statistics, with round and update indices suppressed on within-batch quantities where unambiguous.
The selected frozen reference supplies $\widehat a_{t,g}$ by applying the same statistic to the same records, masks, and temperature.
The resulting group loss is
\begingroup
\small
\begin{equation}
 \widehat\ell_{t,g}
 =-\log\sigma\!\left(
 \beta[\widehat C_{\theta,g}-\widehat a_{t,g}]\right).
 \label{eq:fdpo_group_loss}
\end{equation}
\endgroup
\Cref{eq:fdpo_group_loss} encourages the current aggregate to exceed its reference counterpart, and reduces to \Cref{eq:fdpo_pairwise_dpo} for a singleton group.
Reference selection and the explicit anchor are given in \Cref{eq:fdpo_two_references}.

\vspace{1mm}
\noindent\textbf{Emphasizing Weak Comparisons.}
The aggregate also determines how the loss distributes influence within a group.
For the nonempty set of sampled groups $\mathcal H_t\subseteq\mathcal G_t$, assign equal weights $\rho_g=1/|\mathcal H_t|$.
Differentiating \Cref{eq:fdpo_group_loss} gives
\begingroup
\small
\begin{equation}
\begin{aligned}
 &\frac{\partial(\rho_g\widehat\ell_{t,g})}{\partial s_\theta(j)}\\[2pt]
 &\qquad=-\rho_g\beta\,\omega_{\theta,j}\,
 \sigma\!\left(-\beta\bigl[\widehat C_{\theta,g}-\widehat a_{t,g}\bigr]\right),
\end{aligned}
\label{eq:fdpo_gradient}
\end{equation}
\endgroup
where the normalized gradient weights are
\begingroup
\small
\begin{equation}
 \omega_{\theta,j}
 =\frac{\exp\!\left(-s_\theta(j)/\tau\right)}
 {\displaystyle\sum_{i\in B_{t,g}}\exp\!\left(-s_\theta(i)/\tau\right)}.
 \label{eq:fdpo_gradient_weights}
\end{equation}
\endgroup
The weights in \Cref{eq:fdpo_gradient_weights} sum to one within each group, with group and batch indices implicit.
They measure the sensitivity of $\widehat C_{\theta,g}$ to each margin.
\Cref{eq:fdpo_gradient} gives lower margins greater relative influence, with a common reference-dependent multiplier.
This effect follows from smooth-minimum aggregation~\cite{li2023tiltedlossesmachinelearning}.
The appendix derives the score-space weighting, whose effects can interact under shared parameters.

\subsection{Continual Learning with Class-Routed References}
\label{sec:fdpo_continual}
The reference must distinguish corrections involving established classes from those involving newly introduced ones.
For $t\ge1$, preference training starts from $\pi_t^{\mathrm{SFT}}$, obtained by supervised fine-tuning on current reviewed outputs, while $\pi_{t-1}$ and $\pi_t^{\mathrm{SFT}}$ remain frozen as reference snapshots.
Partition $\mathcal G_t$ into $\mathcal G_t^{\mathrm{old}}$ and $\mathcal G_t^{\mathrm{new}}$ according to the class being corrected.
For label edits, this is the verified class, so an earlier class paired with an incorrect new label still uses the previous model.
Omission and localization groups follow the affected flake's class, while false-positive groups follow the incorrectly assigned class.
We select the reference and compute its matched aggregate as follows,
\begingroup
\small
\begin{equation}
\begin{aligned}
 \pi_{\mathrm{ref},t,g}
 &=\begin{cases}
 \pi_{t-1}, & g\in\mathcal G_t^{\mathrm{old}},\\
 \pi_t^{\mathrm{SFT}}, & g\in\mathcal G_t^{\mathrm{new}},
 \end{cases}\\
 \widehat a_{t,g}
 &=-\tau\log\left[
 \frac{1}{n_g}\sum_{j\in B_{t,g}}
 e^{-s_{\mathrm{ref},t,g}(j)/\tau}\right].
\end{aligned}
\label{eq:fdpo_two_references}
\end{equation}
\endgroup
Here, $s_{\mathrm{ref},t,g}(j)$ uses the selected model in \Cref{eq:fdpo_masked_likelihood,eq:fdpo_score}, and equals $s_{\mathrm{ref},t}(j)$ for a record in group $g$.
\Cref{eq:fdpo_two_references} anchors earlier-class corrections to the previous model and new-class corrections to the post-SFT model, which has already been trained on the introduced classes.

\vspace{1mm}
\noindent\textbf{Training Objective.}
\label{sec:fdpo_training}
We sample groups uniformly and combine their preference losses with supervised grounding:
\begingroup
\small
\begin{equation}
 \mathcal L_t^{\mathrm{train}}
 =\mathcal L_t^{\mathrm{SFT}}
 +\lambda\sum_{g\in\mathcal H_t}\rho_g\widehat\ell_{t,g}.
 \label{eq:fdpo_continual_loss}
\end{equation}
\endgroup
In \Cref{eq:fdpo_continual_loss}, $\lambda>0$ controls preference supervision and $\mathcal L_t^{\mathrm{SFT}}$ includes whole-image characterization and box-conditioned label supervision.
Only the trainable model is updated, and inference directly generates flake boxes and layer labels from the image and instructions.

\subsection{Discrimination and Retention Analysis}
\label{sec:fdpo_fisher_connection}
We now analyze how the group objective relates to discrimination across appearances and retention of earlier preference rankings.
For a fixed group and round, let $Q_g$ be the population distribution of preference records, $J\sim Q_g$, and $S_\theta=s_\theta(J)$.
Assuming $0<\mathbb E_{Q_g}e^{-S_\theta/\tau}<\infty$, define the population aggregate $C_{\theta,Q_g}=-\tau\log\mathbb E_{Q_g}e^{-S_\theta/\tau}$.

\vspace{1mm}
\noindent\textbf{Discrimination Across Appearances.}
If $S_\theta\sim\mathcal N(\mu_g,v_g)$, with score mean $\mu_g$ and variance $v_g$, the aggregate becomes
\begingroup
\small
\begin{equation}
 C_{\theta,Q_g}=\mu_g-\frac{v_g}{2\tau}.
 \label{eq:fdpo_gaussian}
\end{equation}
\endgroup
\Cref{eq:fdpo_gaussian} rewards a stronger average margin while penalizing variation in that margin across observations.
This interpretation motivates the following connection between class separation and within-class appearance variation.

\begin{fdpoproposition}[Fisher Discriminant Connection]
\label{prop:fdpo_fisher}
Let fixed features $Z\in\mathbb R^p$ of distinct material-layer classes $c,c'$ of the same material satisfy $Z\mid c\sim\mathcal N(\mu_c,\Sigma_c)$ and $Z\mid c'\sim\mathcal N(\mu_{c'},\Sigma_{c'})$, where $p$ is the feature dimension and $\mu_c,\Sigma_c$ denote the class mean and covariance.
Assume the score difference between the two labels for the same box is $h(Z)=w^\top Z+b_0$, with $w\in\mathbb R^p$ and a free intercept $b_0$.
The correct margins are $h(Z)$ for class $c$ and $-h(Z)$ for class $c'$.
With equally weighted directions, fixed reference anchors, and positive-definite within-class scatter $S_W=\Sigma_c+\Sigma_{c'}$, minimizing the two-group population preference loss yields
\begingroup
\small
\begin{equation}
 w^\star=\tau S_W^{-1}(\mu_c-\mu_{c'}).
 \label{eq:fdpo_fisher_direction}
\end{equation}
\endgroup
\end{fdpoproposition}

Proposition~\ref{prop:fdpo_fisher} recovers the Fisher discriminant direction~\cite{fisher1936use} in \Cref{eq:fdpo_fisher_direction} directly from the group preference loss.
The solution accounts for both differences between class means and variation within each class. These Gaussian and linear assumptions support the analysis only, while training uses observed margins without fitting feature distributions or estimating covariances.
The appendix provides the proof and extends the score analysis to general distributions.

\vspace{1mm}
\noindent\textbf{Retention Under Comparison Coverage.}
\label{sec:fdpo_retention_goal}
For an earlier group, let $P_g^{\mathrm{old}}$ be the distribution of comparisons to protect, and let $s_{t-1}$ denote their previous-model margins.
Define $F_{t,g}=\Pr_{P_g^{\mathrm{old}}}(s_{t-1}>0,\ s_\theta\le0)$ as the probability that a previously correct preference becomes incorrect or tied.
The following result connects this event to population-group loss when current comparisons cover the protected distribution.

\begin{fdpoproposition}[Retention Under Comparison Coverage]
\label{prop:fdpo_retention}
Suppose $P_g^{\mathrm{old}}\ll Q_g$ and $dP_g^{\mathrm{old}}/dQ_g\le\kappa_g<\infty$ almost surely, where $\ll$ denotes absolute continuity and $\kappa_g$ bounds the density ratio.
Let $a_{t,g}=C_{t-1,Q_g}$ be the finite population aggregate evaluated with $\pi_{t-1}$, and let $\ell_{t,g}=-\log\sigma[\beta(C_{\theta,Q_g}-a_{t,g})]$.
Under the exponential-moment condition above,
\begingroup
\small
\begin{equation}
 F_{t,g}\le\min\!\left\{1,
 \kappa_g e^{-a_{t,g}/\tau}
 (e^{\ell_{t,g}}-1)^{1/(\beta\tau)}\right\}.
 \label{eq:fdpo_forgetting_bound}
\end{equation}
\endgroup
\end{fdpoproposition}

For fixed coverage and reference anchor, reducing the population loss tightens the upper bound in \Cref{eq:fdpo_forgetting_bound}.
Proposition~\ref{prop:fdpo_retention} concerns population quantities $a_{t,g}$ and $\ell_{t,g}$, distinguished from their hatted batch estimates.
Fresh observations of earlier classes can support this coverage assumption without historical-image replay, but a round containing only a new material does not provide coverage of absent materials.
The result bounds preference reversals on covered comparisons, while retention of generated characterizations is evaluated empirically.
\textit{The proof and a comparison with pair-conditioned distillation are provided in the appendix}.
\vspace{-2mm}
\section{Experimental Results}
\label{sec:exp}

\begin{table*}[t]
    \centering
    \caption{\textbf{Material-Incremental Characterization.}
Materials are introduced sequentially in the listed order, with training restricted to the current material and evaluation covering all previously introduced materials.
Material columns report AP50 after the final round, and Final is their equal-weight mean.
MFT, MAA, and BWT measure performance at introduction, average performance across rounds, and backward transfer, respectively.}
    \label{tab:exp_continual_summary_material}
    \footnotesize
    \setlength{\tabcolsep}{3.5pt}
    \begin{tabular*}{\textwidth}{@{\extracolsep{\fill}}l*{9}{r}@{}}
        \toprule
        & \multicolumn{5}{c}{Final AP50}
        & \multicolumn{4}{c}{Continual Performance} \\
        \cmidrule(lr){2-6}\cmidrule(l){7-10}
        Method & Graphite$\uparrow$ & WSe$_2$$\uparrow$
        & WS$_2$$\uparrow$ & MoS$_2$$\uparrow$ & WTe$_2$$\uparrow$
        & Final$\uparrow$ & MFT$\uparrow$ & MAA$\uparrow$ & BWT$\uparrow$ \\
        \midrule
        SFT
        & 17.4 & 0.0 & \textbf{75.2} & 11.5 & 15.8
        & 24.0 & 26.3 & 31.8 & $-2.9$ \\
        EWC~\cite{kirkpatrick2017overcoming}
        & 17.9 & 0.0 & 42.2 & 29.8 & 21.5
        & 22.3 & 30.6 & 33.1 & $-10.4$ \\
        O-LoRA~\cite{wang2023orthogonal}
        & 20.2 & 37.9 & 28.7 & 15.2 & 2.3
        & 20.8 & 20.0 & 29.2 & $+1.1$ \\
        SEFE~\cite{chen2025sefe}
        & 19.1 & 33.7 & 62.9 & \textbf{35.4} & 18.8
        & 34.0 & 27.7 & 35.4 & $\mathbf{+7.8}$ \\
        LwF~\cite{li2017learning}
        & 6.1 & 0.0 & 21.1 & 25.6 & 14.7
        & 13.5 & 21.9 & 25.4 & $-10.4$ \\
        DPO~\cite{rafailov2023direct}
        & 22.3 & 0.0 & 50.5 & 17.8 & 18.5
        & 21.8 & 34.7 & 33.8 & $-16.1$ \\
        $\phi$-DPO~\cite{truong2026faidpo}
        & 22.3 & 0.0 & 56.4 & 11.5 & \textbf{21.9}
        & 22.4 & 36.3 & 30.0 & $-17.3$ \\
        \midrule
        \textbf{DiPACT (Ours)}
        & \textbf{24.6} & \textbf{50.5} & \textbf{75.2} & 32.1 & 13.7
        & \textbf{39.2} & \textbf{36.6} & \textbf{39.1} & $+3.3$ \\
        \bottomrule
    \end{tabular*}
\end{table*}


\begin{figure*}[t]
    \centering
    \includegraphics[width=1\textwidth]{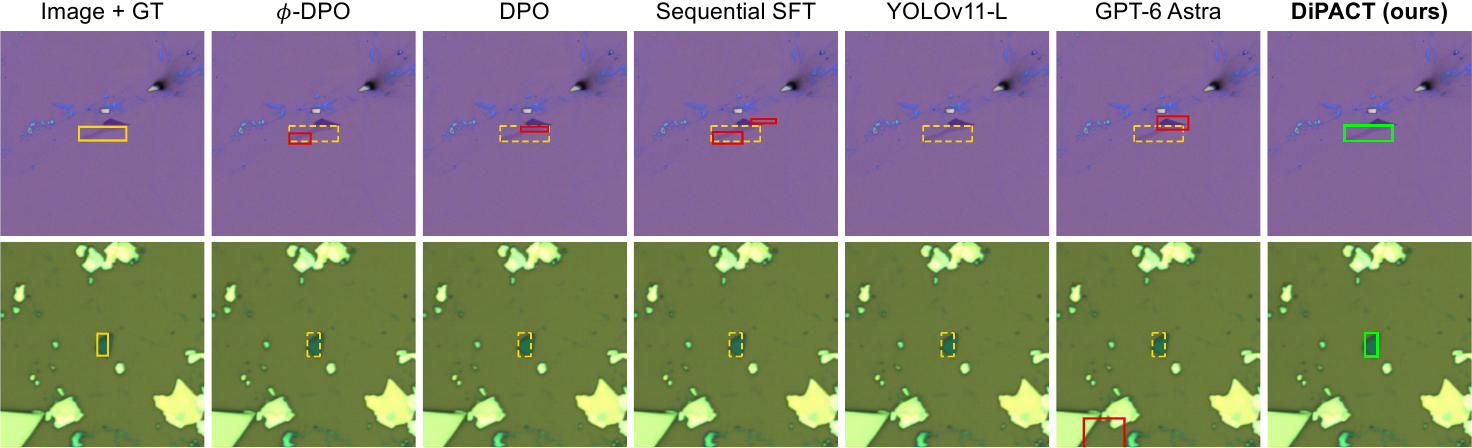}
    \caption{\textbf{Qualitative Characterization.}
    Selected graphite (top) and MoS$_2$ (bottom) examples.
    Yellow boxes mark ground truth, green boxes indicate matched predictions, and red boxes indicate false positives.
    Sequential models are evaluated after the final round, and GPT-6 Astra uses one example per material \textbf{(Best viewed in color)}.}
    \label{fig:exp_qualitative}
\end{figure*}

\subsection{Benchmarks, Metrics, and Implementation}
\label{sec:exp_data}
\label{sec:exp_setup}

\noindent\textbf{Benchmarks and Metrics.}
We evaluate material-incremental learning over graphite, WSe$_2$, WS$_2$, MoS$_2$, and WTe$_2$, and layer-incremental learning over graphite classes 
$1L \rightarrow +2L \rightarrow +3L$.
All microscopy results evaluate thin flakes using AP50~\cite{lin2014microsoft}.
Layer evaluation averages over admitted 1L, 2L, and 3L classes.
Material evaluation combines monolayer and few-layer annotations into one target category and averages AP50 equally across admitted materials. In-house 2L and 3L labels are treated as few-layer.
Following $\phi$-DPO~\cite{truong2026faidpo}, we report final performance (Final), performance immediately after class or material introduction (MFT), average performance across rounds including round 0 (MAA), and average backward transfer on earlier classes or materials (BWT).

\noindent\textbf{Implementation.}
We initialize from QuPAINT-4B~\cite{nguyen2026qupaint} and train rank-16 adapters in the visual encoder, projector, and language decoder using AdamW with learning rate $10^{-4}$.
The shared initial round uses 1,500 SFT steps.
Layer-incremental updates use 300 SFT steps followed by 700 method-specific steps.
Material-incremental training schedules are reported in the supplementary material.
The layer-incremental comparisons use the same backbone, data access, and training budget.
All preference methods use the same reference construction within each benchmark, with $\beta=0.5$ and preference weight $\lambda=0.5$.
DiPACT uses temperature $\tau=1$, and $\phi$-DPO uses focal exponent $\gamma=2$.

\subsection{Main Results}
\label{sec:exp_main}

\noindent\textbf{Material-Incremental Learning.}
\label{sec:exp_material}
In \Cref{tab:exp_continual_summary_material},
DiPACT achieves the best final and sequence-wide performance,
improving Final AP50 and MAA over the strongest baselines by
5.2 and 3.7 points, respectively.
DiPACT also achieves the highest MFT (36.6), slightly
exceeding $\phi$-DPO (36.3), while substantially improving
final AP50 from 22.4 to 39.2.
The lower WTe$_2$ score may reflect a trade-off between preserving earlier distinctions and adapting to the newest material.
Overall, these results support effective acquisition alongside stronger characterization across the accumulated material set.

\noindent\textbf{Comparison with General Models and Detectors.}
\label{sec:exp_general}
\Cref{tab:exp_model_comparison} compares final characterization across five materials.
DiPACT reaches 39.2 AP50, compared with 24.0 for sequential SFT, 18.4 for sequential YOLOv11-L, and at most 12.1 for the evaluated frozen general models.
Qwen3-VL-4B~\cite{yang2025qwen3} and GPT-6 Astra~\cite{openai2026gpt6astra} use zero-shot and one-shot prompts. YOLO uses 40 epochs. These results highlight the limitations of off-the-shelf
vision-language models for microscopy
characterization, even with an in-context example,
and support the need for task-specific adaptation.

\begin{table}[t]
    \centering
    \caption{AP50 averaged equally across all five materials
    and the four with at least five annotated targets in the evaluation set. Sequential models are evaluated after the final round.}
    \label{tab:exp_model_comparison}
    \footnotesize
    \setlength{\tabcolsep}{3pt}
    \begin{tabular*}{\linewidth}{@{\extracolsep{\fill}}lrr@{}}
        \toprule
        Model & \shortstack{All\\Materials$\uparrow$} & \shortstack{$\geq5$\\Targets$\uparrow$} \\
        \midrule
        Qwen3-VL-4B (\textit{Zero-Shot}) & 0.7 & 0.8 \\
        Qwen3-VL-4B (\textit{One-Shot}) & 0.0 & 0.0 \\
        GPT-6 Astra (\textit{Zero-Shot}) & 12.1 & 4.5 \\
        GPT-6 Astra (\textit{One-Shot}) & 10.8 & 13.4 \\
        \midrule
        YOLOv11-L (40 Epochs) & 18.4 & 19.8 \\
        QuPAINT & 24.0 & 11.2 \\
        \midrule
        \textbf{DiPACT (Ours)} & \textbf{39.2} & \textbf{30.2} \\
        \bottomrule
    \end{tabular*}
\end{table}

\noindent\textbf{Layer-Incremental Learning.}
\label{sec:exp_layer}
\Cref{tab:exp_continual_summary_layer} evaluates characterization as bilayer and trilayer graphite are introduced.
DiPACT achieves the highest final AP50 of 24.5, exceeding SFT, DPO, and $\phi$-DPO by 1.4, 5.0, and 8.4 points, respectively.
It also achieves the highest MFT of 21.2 and BWT of $+4.8$ among the compared methods. DiPACT's positive BWT, compared with negative BWT for DPO and $\phi$-DPO, is consistent with its emphasis on weak comparisons and reference alignment supporting both new-class learning and retention.

\begin{table}[t]
    \centering
    \caption{\textbf{Layer-Incremental Characterization.} BWT is in AP50 points. All methods share the round-0 initialization.}
    \label{tab:exp_continual_summary_layer}
    \footnotesize
    \setlength{\tabcolsep}{3pt}
    \begin{tabular*}{\linewidth}{@{\extracolsep{\fill}}lrrrr@{}}
        \toprule
        Method & Final$\uparrow$ & MFT$\uparrow$ & MAA$\uparrow$ & BWT$\uparrow$ \\
        \midrule
        SFT & 23.1 & 20.7 & 18.5 & $+3.7$ \\
        LwF & 13.7 & 12.6 & 15.9 & $+1.7$ \\
        DPO~\cite{rafailov2023direct} & 19.5 & 21.1 & 18.7 & $-2.4$ \\
        $\phi$-DPO~\cite{truong2026faidpo} & 16.1 & 19.0 & \textbf{18.8} & $-4.3$ \\
        \midrule
        \textbf{DiPACT (Ours)} & \textbf{24.5} & \textbf{21.2} & 17.7 & $\mathbf{+4.8}$ \\
        \bottomrule
    \end{tabular*}
\end{table}

\noindent\textbf{Generalization Across Visual Domains.}
To test whether DiPACT's benefits extend beyond material characterization, we apply it to building damage assessment in xBD satellite imagery~\cite{gupta2019xbd}.
Using the general InternVL3.5-4B backbone, we introduce major damage after learning no-damage and destroyed classes, with fresh images and no replay.
In \Cref{tab:exp_xbd_damage}, DiPACT achieves the highest overall damage F1 of 90.0 and new-class F1 of 84.0.
Its damage F1 exceeds SFT, DPO, and $\phi$-DPO by 7.5, 11.5, and 2.3 points, respectively.

\begin{table}[t]
    \centering
    \caption{\textbf{Cross-Domain Evaluation on xBD Satellite Imagery.}
    Results after adding major damage, on 300 test tiles with 1,759 admitted-class buildings.
    F1 scores use buildings matched at IoU $\geq0.5$. Damage F1 is the harmonic mean of the three class F1 scores.}
    \label{tab:exp_xbd_damage}
    \footnotesize
    \setlength{\tabcolsep}{3pt}
    \begin{tabular*}{\linewidth}{@{\extracolsep{\fill}}lrr@{}}
        \toprule
        Method & Damage F1$\uparrow$ & \shortstack{Major-damage F1$\uparrow$\\(new class)} \\
        \midrule
        SFT & 82.5 & 72.7 \\
        DPO~\cite{rafailov2023direct} & 78.5 & 60.6 \\
        $\phi$-DPO~\cite{truong2026faidpo} & 87.7 & 80.0 \\
        \midrule
        \textbf{DiPACT (Ours)} & \textbf{90.0} & \textbf{84.0} \\
        \bottomrule
    \end{tabular*}
\end{table}

\vspace{1mm}
\noindent\textbf{Qualitative Results.}
\Cref{fig:exp_qualitative} compares predictions on graphite and MoS$_2$ images.
In these examples, DiPACT localizes the annotated flakes, while the compared methods miss them or produce false positives.

\providecommand{\ablYes}{\textcolor{green!50!black}{$\checkmark$}}
\providecommand{\ablNo}{\textcolor{red!75!black}{$\times$}}

\subsection{Ablation Study}
\label{sec:exp_ablation}

\noindent\textbf{Component Ablations.}
\Cref{tab:exp_ablation} compares aggregation, correction types, and reference selection during round 2 from a shared round-1 checkpoint, using the earlier whole-answer scoring configuration.
The full method achieves the highest final AP50, MAA, and BWT among these variants.
Mean aggregation and label edits alone yield final AP50 of 20.7 and 19.6, respectively, compared with 21.1 for the full method.
Using only the previous or post-SFT reference yields 19.0 or 17.9 AP50 and negative BWT, compared with $+3.4$ for routed references. 

\noindent\textbf{Effect of Temperature.}
\Cref{tab:exp_tau} varies $\tau$ from a common post-SFT
checkpoint using 350 preference-stage updates in round 1.
$\tau=1$ achieves the highest AP50, MFT, and MAA,
while increasing $\tau$ to 4 reduces AP50 from 20.7 to 16.0.
The highest BWT occurs at $\tau=2$, indicating that stronger
backward transfer alone does not ensure better final characterization.

\begin{table}[t]
    \centering
    \caption{\textbf{Component Ablations and Temperature Sensitivity.}
    Components use round-2 whole-answer scoring.
    Temperature uses round-1 final scoring with 350 updates. BWT is in AP50 points. SM: smooth minimum. DP: detection preferences.}
    \label{tab:exp_ablation}
    \label{tab:exp_tau}
    \vspace{-2mm}
    \footnotesize
    \setlength{\tabcolsep}{2pt}
    \begin{tabular*}{\linewidth}{@{\extracolsep{\fill}}ccccrrrr@{}}
        \toprule
        \multicolumn{4}{c}{Components} & \multicolumn{4}{c}{Performance} \\
        \cmidrule(lr){1-4}\cmidrule(l){5-8}
        SM & DP & \shortstack{Prev.\\Ref.} & \shortstack{Post-SFT\\Ref.}
        & AP50$\uparrow$ & MFT$\uparrow$ & MAA$\uparrow$ & BWT$\uparrow$ \\
        \midrule
\ablNo  & \ablYes & \ablYes & \ablYes
        & 20.7 & 22.8 & 19.5 & $-3.1$ \\
        \ablYes & \ablNo  & \ablYes & \ablYes
        & 19.6 & 23.5 & 19.1 & $-5.7$ \\
        \ablYes & \ablYes & \ablYes & \ablNo
        & 19.0 & 20.8 & 18.9 & $-2.8$ \\
        \ablYes & \ablYes & \ablNo  & \ablYes
        & 17.9 & \textbf{23.7} & 18.5 & $-8.8$ \\
        \midrule
        \ablYes & \ablYes & \ablYes & \ablYes
        & \textbf{21.1} & 18.8 & \textbf{19.6} & $\mathbf{+3.4}$ \\
        \bottomrule
    \end{tabular*}
    \par\vspace{1mm}
    \begin{tabular*}{\linewidth}{@{\extracolsep{\fill}}lrrrr@{}}
        Temperature $\tau$ & AP50$\uparrow$ & MFT$\uparrow$ & MAA$\uparrow$ & BWT$\uparrow$ \\
        \midrule
0.25 & 18.4 & 15.2 & 15.8 & $+6.3$ \\
        0.5 & 18.4 & 15.9 & 15.7 & $+5.0$ \\
        1.0 & \textbf{20.7} & \textbf{19.3} & \textbf{16.9} & $+2.9$ \\
        2.0 & 20.3 & 15.5 & 16.7 & $\mathbf{+9.6}$ \\
        4.0 & 16.0 & 14.7 & 14.6 & $+2.7$ \\
        \bottomrule
    \end{tabular*}
    \vspace{-2mm}
\end{table}

\vspace{-2mm}
\section{Conclusion}
\label{sec:conclusion}
We have introduced DiPACT for continual grounded material characterization.
Its group-level preference objective has emphasized weak comparisons across images, while reference alignment has guided new-class acquisition and knowledge retention.
We have connected the objective to Fisher discrimination under linear-Gaussian assumptions, established a conditional preference-retention bound, and introduced CoMAt with layer- and material-incremental protocols.
Experiments have demonstrated state-of-the-art continual characterization performance among the evaluated methods and benefits beyond the materials domain.

\noindent\textbf{Limitations.}
The Fisher analysis relied on linear-Gaussian assumptions, and the retention bound required coverage of earlier comparisons.
The grounded VLM is still limited by the maximum token length, which can make it difficult to handle dense scenes.


\section*{Acknowledgments} 
This work is partly supported by MonArk NSF Quantum Foundry (DMR-1906383), NSF PCL-Test Bed AURORA (2607535) and  NSF Quantum Award (2444042). It acknowledges the Arkansas High-Performance Computing Center for providing GPUs.
{
    \small
    \bibliographystyle{ieeenat_fullname}
    \bibliography{main}
}


\end{document}